\documentclass[a4paper,fleqn]{cas-dc}
\usepackage[numbers,sort&compress]{natbib}
\usepackage{multirow}
\usepackage{graphicx}
\usepackage{amsmath,amssymb}
\usepackage{booktabs}
\usepackage[T1]{fontenc}
\usepackage[utf8]{inputenc}
\usepackage[expansion=false]{microtype} 
\begin{document}
\let\WriteBookmarks\relax
\shorttitle{RCT-AD: Reliability-Aware Temporal BEV Reasoning}
\shortauthors{A. Dey et al.}
\title[mode=title]{RCT-AD: Reliability-Aware Context and Temporal BEV Reasoning for End-to-End Autonomous Driving}
\author[1]{Argho Dey}[orcid=0009-0000-6739-9632]
\ead{arghodey@stu.cqu.edu.cn}
\author[1]{Yunfei Yin}[orcid=0000-0003-1792-1378]
\cormark[1]
\ead{yinyunfei@cqu.edu.cn}
\author[1]{Swachha Ray}[orcid=0009-0009-6900-9859]
\ead{swachharay@stu.cqu.edu.cn}
\author[1]{Md Minhazul Islam}[orcid=0009-0000-8423-5066]
\ead{mishaown@stu.cqu.edu.cn}
\author[1]{Zheng Yuan}[orcid=0009-0005-8454-4383]
\ead{yuanzheng@cqu.edu.cn}
\author[1]{Sijing Xiong}[orcid=0009-0006-1268-5272]
\ead{202414021016t@stu.cqu.edu.cn}
\author[1]{Hongyu Liu}[orcid=0009-0009-6189-8970]
\ead{202514131048T@stu.cqu.edu.cn}
\author[1]{Zhiqiu Huang}[orcid=0009-0005-5568-0749]
\ead{202514131200@stu.edu.cqu.cn}
\affiliation[1]{organization={College of Computer Science, Chongqing University},
 city={Chongqing}, postcode={400044}, country={China}}
\cortext[1]{Corresponding author.}
\begin{abstract}
Reliable temporal representation learning is important for end-to-end autonomous driving, particularly when visual observations are degraded by occlusion, motion blur, illumination variation, and sensor noise. Existing Bird’s-Eye-View (BEV) approaches provide a unified representation for perception and planning; however, temporal features are commonly aggregated across frames without explicitly accounting for the reliability of individual observations. Consequently, low-quality or inconsistent features may propagate through temporal representations and degrade downstream perception, prediction, and planning. To address this problem, we propose RCT-AD, a Reliability-Aware Context and Temporal BEV Reasoning framework for end-to-end autonomous driving that explicitly models feature reliability and temporal consistency within a shared BEV representation. RCT-AD introduces a Reliable Context Awareness (RCA) module that estimates frame-level reliability using complementary spatial, semantic, temporal, and visual-quality cues, and selectively preserves trustworthy information through a reliability-guided temporal memory mechanism. When the current observation is unreliable, historical high-reliability features are aligned and integrated to construct a more stable temporal BEV representation. A Temporal Trajectory Planner further models long-term temporal dependencies and multi-agent interactions, while semantic segmentation and motion-guided detection provide complementary supervision for refining the shared BEV representation. Experiments on the nuScenes benchmark \cite{caesar2020nuscenes} demonstrate consistent improvements across perception, motion prediction, and trajectory planning. RCT-AD achieves 61.5 NDS, 52.9 mAP, and 52.3 mIoU while maintaining competitive computational efficiency. These results demonstrate the effectiveness of reliability-aware temporal BEV reasoning for unified perception, prediction, and planning. 

\end{abstract}
\maketitle

\section{Introduction}

Autonomous driving has emerged as an important research area in artificial intelligence \cite{chen2024endtoend}, requiring the integration of visual perception, scene understanding, motion prediction, and decision-making in complex and dynamically changing environments. Recent advances in deep learning, multi-view perception, and representation learning have substantially improved the ability of autonomous systems to construct structured scene representations and predict future motion. Nevertheless, many autonomous driving systems still adopt modular pipelines in which perception \cite{wang2023objectcentric}, prediction \cite{valiente2024prediction}, \cite{zhou2023querycentric}, and planning \cite{hu2023planning}, \cite{wang2024driving} are optimized as separate components. Although such modular designs provide interpretability and facilitate independent optimization, they can interrupt information flow across different tasks. Consequently, errors introduced during perception may propagate to subsequent prediction and planning stages, particularly in uncertain and dynamically changing environments \cite{kendall2017uncertainty}, \cite{lakshminarayanan2017simple}.

Recent end-to-end autonomous driving frameworks attempt to alleviate this limitation by jointly learning perception, prediction, and planning within a unified architecture. In particular, Bird’s-Eye-View (BEV) representations provide a shared spatial feature space in which multi-view observations can be jointly exploited for object detection, semantic understanding, motion prediction, and trajectory planning. Methods such as BridgeAD \cite{zhang2025bridging} further connect historical perception with future prediction and planning through shared representations. However, a key challenge remains in the reliability of temporal representation learning. Many BEV-based approaches \cite{li2022bevformer}, \cite{huang2021bevdet}, \cite{li2023bevdepth}, \cite{gu2021belief} aggregate information across multiple frames to improve temporal consistency, but they typically do not explicitly evaluate whether individual observations are sufficiently reliable before being propagated through the temporal representation. In practical scenarios, visual observations may be degraded by occlusion, motion blur, illumination variation, or sensor noise. Directly incorporating such unreliable observations can introduce corrupted or inconsistent features into the shared BEV space, allowing errors to accumulate over time and adversely affect multiple downstream tasks. Therefore, effective temporal BEV reasoning should consider not only how historical information is aggregated, but also which observations are sufficiently reliable to preserve and propagate.

Another challenge concerns the effective integration of semantic context and long-term temporal information within a unified BEV representation. Detection-oriented BEV representations primarily emphasize object-level geometric information and may provide insufficient contextual knowledge about drivable regions, lane structures, scene layout, and interactions among traffic participants. Such semantic information is important for learning structured scene representations that support subsequent prediction and planning. Meanwhile, temporal reasoning in existing frameworks is frequently based on fixed-range feature aggregation or short-term historical information, which may limit the modeling of longer-term motion dependencies and multi-agent interactions. Therefore, a unified framework should not only exploit historical observations, but also determine which temporal information is sufficiently reliable to preserve and propagate, while maintaining semantic and interaction-aware representations for downstream tasks.

\begin{figure*}[!t]
\centering
\includegraphics[width=0.85\textwidth]{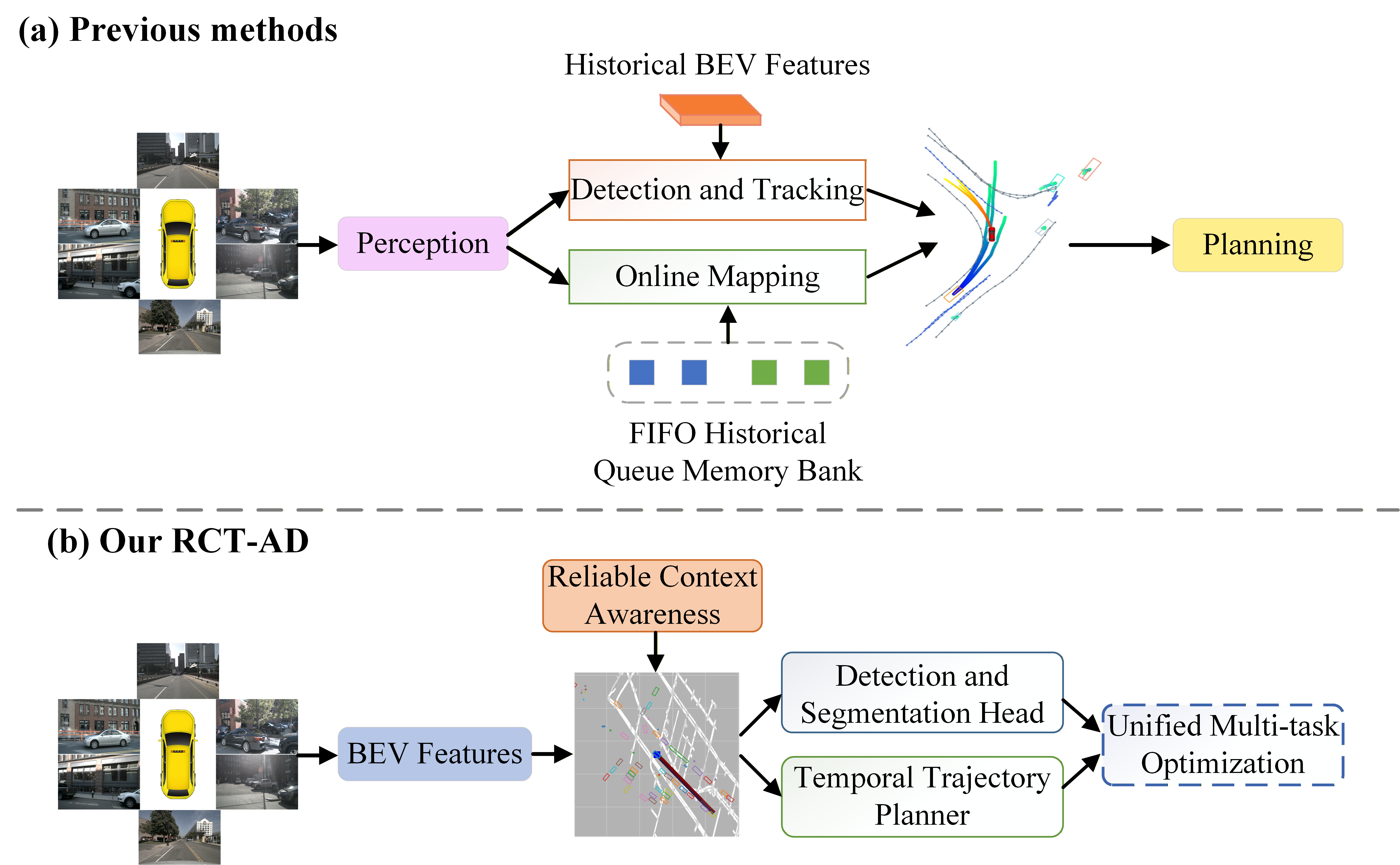}
\caption{Motivation of RCT-AD. Conventional temporal autonomous-driving
frameworks propagate historical BEV features without explicitly assessing
their reliability, whereas RCT-AD performs reliability-aware temporal
feature selection and refinement before downstream perception and planning.}
\label{fig:motivation}
\end{figure*}

To address these challenges, we propose RCT-AD, a Reliability-aware Context and Temporal BEV reasoning framework for end-to-end autonomous driving. RCT-AD formulates temporal BEV aggregation from a reliability-aware representation-learning perspective by explicitly evaluating the quality of observations before historical information is propagated through the shared feature space. Specifically, the proposed Reliable Context Awareness (RCA) mechanism estimates the reliability of BEV features using complementary spatial, confidence, semantic-uncertainty, temporal-stability, and visual-quality cues. Reliable observations are selectively preserved in a temporal memory bank, whereas degraded observations are refined using aligned high-reliability historical context. This mechanism reduces the influence of unreliable temporal information and maintains more consistent BEV representations across consecutive frames. On top of the reliability-refined representation, a Temporal Trajectory Planner captures long-term dependencies and interactions among agents, while semantic segmentation and motion-guided detection provide complementary supervision for improving the structural representation of the driving scene. In this way, RCT-AD integrates reliability-aware temporal reasoning, semantic representation learning, and multi-task prediction within a unified BEV framework.

Our main contributions are as follows:

\begin{itemize}

\item \textbf{Reliability-aware temporal BEV representation.}
We introduce a Reliable Context Awareness (RCA) mechanism that explicitly evaluates the reliability of BEV observations using multiple complementary quality cues and performs reliability-guided temporal feature routing. A reliability-aware memory mechanism selectively preserves trustworthy historical information and uses high-reliability context to refine degraded observations, thereby reducing the propagation of unreliable features and improving temporal representation consistency.

\item \textbf{Reliability-guided temporal trajectory reasoning.}
We develop a Temporal Trajectory Planner that operates on reliability-refined BEV representations and integrates recurrent temporal encoding with graph-based interaction reasoning. By exploiting historical context together with agent-to-agent and agent-to-map interactions, the planner models long-term motion dependencies and generates temporally consistent trajectory predictions for both the ego vehicle and surrounding agents.

\item \textbf{Semantic and motion-guided BEV refinement.}
We incorporate joint detection and segmentation supervision to introduce semantic structure and motion information into the shared BEV representation. This multi-task learning strategy enhances scene-level contextual representation and provides complementary perceptual information for downstream prediction and planning.

\end{itemize}

\section{Related Work}
\subsection{Semantic-Guided Learning and Scene Understanding}

Semantic-guided learning~\cite{hu2023planning,wang2024driving} plays an important role in
enabling autonomous systems to understand driving environments beyond
purely geometric information. By incorporating semantic cues such as
drivable regions, lane structures, and obstacle categories, learned
representations can provide richer contextual information for downstream
prediction and planning. Early approaches~\cite{pan2020crossview} primarily focused
on cross-view semantic segmentation or BEV semantic mapping
~\cite{li2022bevformer,philion2020lift} to improve scene perception. However, semantic
information was often treated as an auxiliary prediction target rather
than being explicitly incorporated into the shared representation used
for decision-making. 

More recent end-to-end frameworks~\cite{zhang2025bridging,zheng2024genad} integrate semantic
information directly into feature representations or planning modules,
allowing geometric and contextual information to jointly influence
downstream predictions. Such semantic-aware representations provide
structured information about drivable space, scene layout, and traffic
participants, thereby improving the consistency between visual perception
and trajectory reasoning in complex driving environments.

\subsection{Reliability-Aware Temporal Modeling and Memory-Gated Learning in Autonomous Driving}

Temporal modeling and memory mechanisms~\cite{li2022bevformer,wang2023objectcentric} are important
for autonomous systems to capture scene evolution, object motion, and
long-term dependencies across consecutive observations. In contrast to
single-frame perception~\cite{liu2022petr,harley2023simplebev}, temporal reasoning enables
learned representations to preserve motion continuity and exploit
historical information when current observations are incomplete or
ambiguous. Existing approaches have employed multi-frame feature
aggregation, temporal attention, recurrent networks, and object-centric
memory mechanisms to improve temporal consistency. 

However, most temporal aggregation strategies primarily focus on how
historical information is incorporated, while the reliability of individual
observations is not always explicitly considered. Visual degradation caused
by occlusion, motion blur, illumination variation, or sensor noise may
introduce unreliable features into temporal representations. Once propagated
through memory, such features can influence subsequent perception,
prediction, and planning. Uncertainty-aware and robustness-oriented learning
methods~\cite{kendall2017uncertainty,lakshminarayanan2017simple,meyer2020uncertainty,song2024robustness} have investigated confidence estimation
and robustness in visual perception, but explicit reliability-guided
selection and reconstruction of temporal BEV features remain comparatively
underexplored. This motivates adaptive temporal memory mechanisms that
selectively preserve trustworthy observations while limiting the propagation
of degraded information. 

\subsection{Unified BEV Representation for Perception and Planning}

Bird's-Eye-View (BEV) representations~\cite{hu2023planning,philion2020lift} provide a structured
spatial feature space for integrating perception, prediction, and planning
within modern autonomous driving systems. By projecting features extracted
from multiple camera views~\cite{li2022bevformer,liu2022petr,zeng2025resar} into a common coordinate
system, BEV representations facilitate consistent reasoning about objects,
scene geometry, mapping information, and future trajectories. 

Early BEV approaches~\cite{li2022bevformer,philion2020lift,huang2021bevdet} mainly focused on perception,
whereas more recent end-to-end frameworks have extended shared BEV
representations to motion prediction and planning. Nevertheless, several
methods~\cite{sun2025sparsedrive,wang2024driving} still rely on loosely coupled task-specific
representations or temporal aggregation mechanisms that do not explicitly
account for observation reliability. This can limit the consistency of
spatial and temporal information propagated across tasks. A unified BEV
representation that combines semantic supervision, temporal reasoning, and
reliability-aware feature propagation therefore provides a promising basis
for robust multi-task autonomous driving.

\section{Methodology}

RCT-AD is a reliability-aware end-to-end autonomous driving framework
that integrates perception, temporal reasoning, motion prediction, and
trajectory planning within a unified Bird's-Eye-View (BEV)
representation. The central objective is to prevent unreliable visual
observations from being indiscriminately propagated through temporal
representations. As illustrated in Fig.~\ref{fig:framework}, multi-view
camera images are first encoded into a shared BEV feature space. The
resulting representation is subsequently processed by the Reliable
Context Awareness (RCA) module, which evaluates observation reliability
and selectively preserves or reconstructs temporal features using
historical context. The reliability-refined BEV representation is then
shared by the perception and planning branches. A Temporal Trajectory
Planner captures long-term motion dependencies and agent interactions,
while semantic and motion-guided refinement provides complementary
scene-level supervision. All components are optimized jointly through a
unified multi-task objective.

\subsection{Framework Overview and Multi-view BEV Feature Encoder}

\begin{figure*}[!t]
    \centering
    \includegraphics[width=0.95\textwidth]{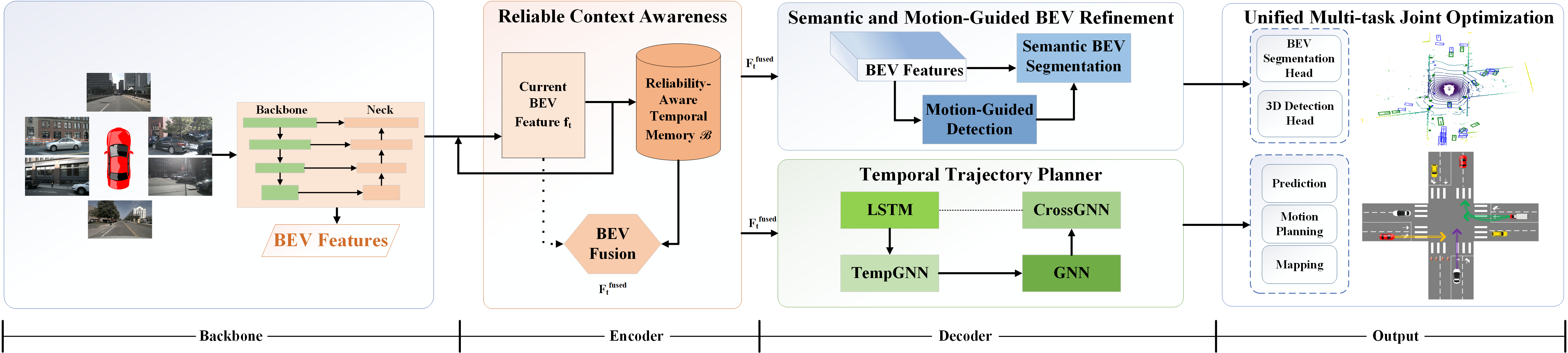}
    \caption{Overall architecture of the proposed RCT-AD framework.
    Multi-view camera observations are first encoded into a shared BEV
    representation. Reliable Context Awareness (RCA) evaluates observation
    reliability and performs reliability-guided temporal feature reconstruction
    and memory fusion. The refined BEV representation is subsequently shared
    by the Temporal Trajectory Planner and the semantic and motion-guided
    perception branches for unified detection, segmentation, motion prediction,
    mapping, and trajectory planning.}
    \label{fig:framework}
\end{figure*}

The overall architecture of RCT-AD is illustrated in
Fig.~\ref{fig:framework}. Given synchronized images captured by $N$
surround-view cameras at time $t$, denoted by
$\mathcal{I}_t=\{I_t^{i}\}_{i=1}^{N}$, RCT-AD first extracts multi-scale
visual features using a shared convolutional backbone. A deformable~\cite{zhu2021deformable} feature neck subsequently aggregates multi-level information, and the
resulting multi-view features are projected into a common BEV coordinate
system. The BEV encoding process can be expressed as

\begin{equation}
F_{t}^{\mathrm{BEV}}
=
\operatorname{Proj}_{\mathrm{BEV}}
\left(
\operatorname{Neck}
\left(
\operatorname{Backbone}
\left(
\{I_t^{i}\}_{i=1}^{N}
\right)
\right)
\right)
\label{eq:bev_encoder}
\end{equation}

where $F_{t}^{\mathrm{BEV}}$ denotes the shared BEV representation at
time $t$. In our implementation, ResNet-50~\cite{he2016resnet} is adopted for RCT-AD-T,
whereas ResNet-101 is used for RCT-AD-L. The shared BEV representation
provides a common latent space for detection, semantic segmentation,
motion prediction, mapping, and trajectory planning, thereby allowing
spatial, semantic, and temporal information to be jointly optimized
across tasks. 

\subsection{Reliable Context Awareness}

Temporal BEV aggregation can improve scene understanding by incorporating
historical information; however, it may also propagate corrupted features
when individual observations are unreliable. To address this problem,
RCT-AD introduces Reliable Context Awareness (RCA), which explicitly
evaluates the quality of each incoming BEV representation before temporal
information is propagated.

RCA consists of three operations: reliability estimation,
reliability-guided feature reconstruction, and reliability-aware memory
fusion. For each incoming frame, RCA first estimates a reliability score
from complementary spatial, confidence, semantic-uncertainty, temporal,
and visual-quality cues. Reliable observations are directly preserved in
temporal memory, whereas low-reliability observations are refined using
aligned historical features. The resulting representations are then
aggregated through reliability-weighted memory fusion to produce the
final temporally refined BEV feature.

\begin{figure}
    \centering
    \includegraphics[width=.9\columnwidth, keepaspectratio]{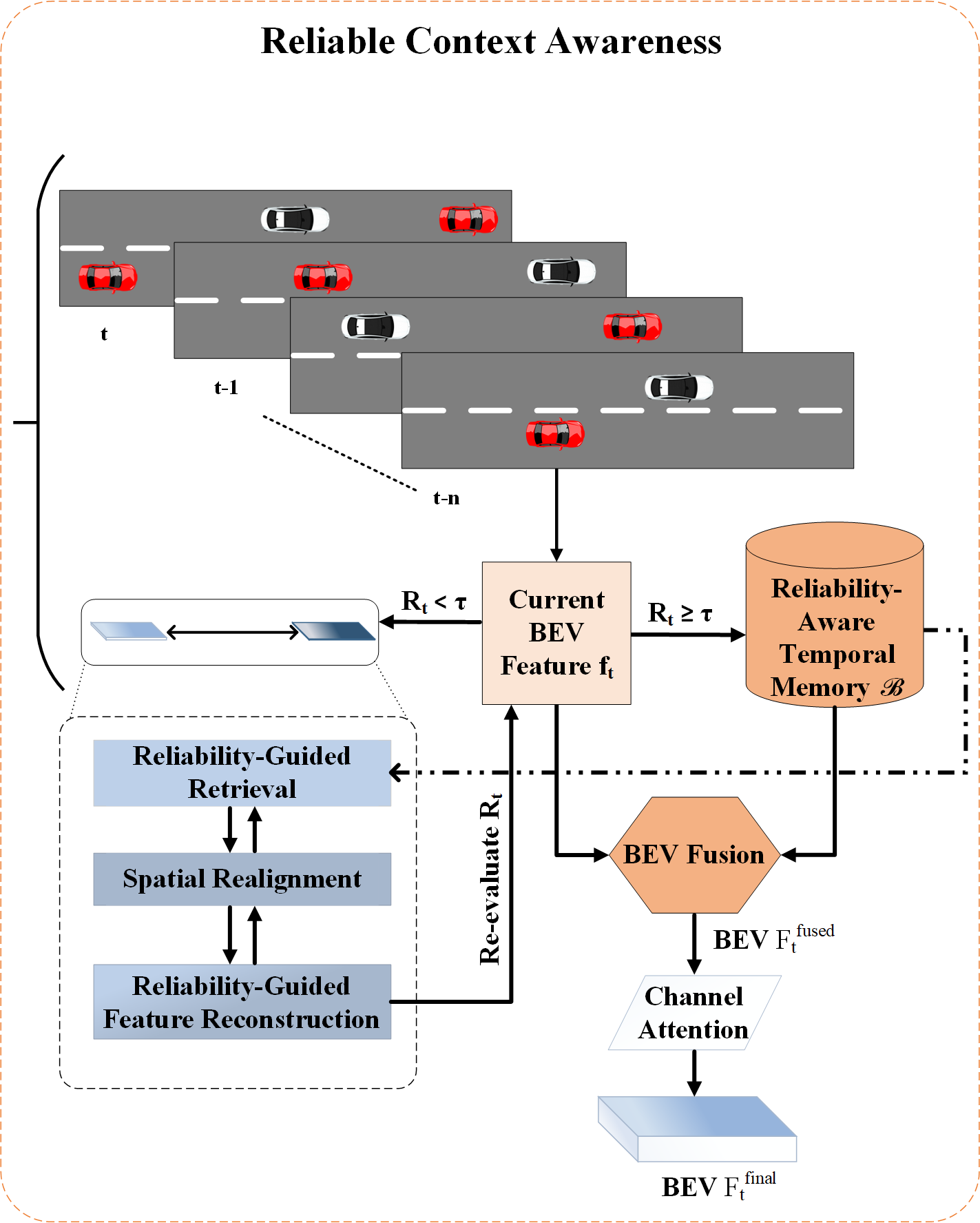}
    \caption{Overview of the proposed Reliable Context Awareness (RCA)
    mechanism. RCA estimates the reliability $R_t$ of the current BEV
    feature $f_t$. When $R_t<\tau$, high-reliability historical features
    are retrieved from the temporal memory, spatially aligned, and used
    for reliability-guided feature reconstruction, after which reliability
    is re-evaluated. When $R_t\geq\tau$, the current feature is preserved
    in the reliability-aware temporal memory $\mathcal{B}$. The historical
    representations are aggregated through reliability-weighted BEV fusion,
    and channel-attention refinement produces the final representation
    $F_t^{\mathrm{final}}$.}
    \label{fig:rca}
\end{figure}

\subsubsection{Reliability Estimation}

For each time step $t$, the BEV encoder produces a feature representation
$f_t$. RCA estimates its reliability using five complementary indicators:

\begin{equation}
R_t =
\alpha_1 \operatorname{IoU}_t
+\alpha_2 \operatorname{Conf}_t
+\alpha_3 (1-H_t)
+\alpha_4 S_t
+\alpha_5 P_t
\label{eq:reliability}
\end{equation}

where $\operatorname{IoU}_t$ measures spatial consistency with the
temporally aligned preceding BEV representation,
$\operatorname{Conf}_t$ denotes the mean detection confidence,
$H_t$ represents normalized segmentation entropy,
$S_t$ measures instance-level temporal stability, and
$P_t$ characterizes visual quality based on image exposure and blur
statistics. Each component is normalized to $[0,1]$, and the
coefficients satisfy

\begin{equation}
\sum_{j=1}^{5}\alpha_j = 1,
\qquad
\alpha_j \geq 0,
\label{eq:alpha_constraint}
\end{equation}

such that $R_t\in[0,1]$.

A reliability threshold $\tau$ determines the subsequent feature-routing
strategy. If $R_t\geq\tau$, the current observation is considered
sufficiently reliable and is inserted into the reliable temporal memory.
Otherwise, it is processed by the reliability-guided reconstruction
procedure described below. 

\subsubsection{Reliability-Guided Feature Reconstruction}

When $R_t < \tau$, RCA refines the current BEV representation using
reliable historical context. Let $f_t^{(0)}=f_t$ denote the initial
low-reliability feature and $R_t^{(0)}=R_t$ its reliability score. At
iteration $k$, the most reliable historical representation is retrieved
from the memory bank $\mathcal{B}$ as

\begin{equation}
m^{*(k)}
=
\underset{m\in\mathcal{B}}{\operatorname{arg\,max}}\; R_m .
\label{eq:memory_retrieval}
\end{equation}

The retrieved historical feature is first aligned with the current
ego-pose and then adaptively combined with the current representation:

\begin{equation}
f_t^{(k)}
=
\lambda^{(k)} f_t^{(k-1)}
+
\left(1-\lambda^{(k)}\right)
\operatorname{Warp}
\left(
f_{m^{*(k)}},
\Delta\mathrm{pose}
\right)
\label{eq:feature_reconstruction}
\end{equation}

where

\begin{equation}
\lambda^{(k)}
=
\frac{R_t^{(k-1)}}
{R_t^{(k-1)}+R_{m^{*(k)}}}.
\label{eq:reconstruction_weight}
\end{equation}

After each reconstruction step, the reliability score is recomputed using
Eq.~\eqref{eq:reliability}. The iterative reconstruction terminates when
$R_t^{(k)}\geq\tau$ or when the maximum number of iterations $K_r$ is
reached. If the reliability threshold is not reached within $K_r$
iterations, RCA uses the pose-aligned highest-reliability historical
representation as a fallback. Bounding the number of reconstruction
iterations limits the additional computational cost and guarantees
termination.

\subsubsection{Reliability-Aware Memory Fusion}

The temporal memory stores accepted representations together with their
reliability and age information. Each entry is represented as
$(f_i,R_i,a_i)$, where $f_i$ denotes the stored feature, $R_i$ its
reliability score, and $a_i$ its temporal age. To regulate the retention
of outdated observations in temporal memory, an age-dependent retention
factor is defined as

\begin{equation}
w_i = \exp(-\beta a_i),
\label{eq:age_decay}
\end{equation}

where $\beta$ controls the temporal decay rate. The age-dependent factor
$w_i$ is used exclusively for temporal-memory management, i.e., to
prioritize the retention or eviction of stored representations according
to their temporal freshness. It does not directly participate in the
subsequent feature-fusion operation. When the memory reaches its capacity,
entries with low reliability or excessive temporal age are preferentially
removed, whereas recent high-reliability representations are retained for
subsequent temporal reasoning. The feature-fusion weights are instead
determined by the reliability scores, as defined in
Eq.~\eqref{eq:reliability_fusion}.

For long-term temporal consistency, each tracked instance $i$ maintains
a representation $h_t^i$ that is updated according to

\begin{equation}
h_{t+1}^{i}
=
\gamma_t^{i} h_t^{i}
+
\left(1-\gamma_t^{i}\right)f_t^{i},
\qquad
\gamma_t^{i}=g(R_t^{i})
\label{eq:memory_update}
\end{equation}

where $g(\cdot)$ is a learnable reliability-gating function implemented
using a sigmoid-activated multilayer perceptron. The gate determines the
relative contribution of historical and current information according to
the estimated reliability.

At decoding time, the stored representations are aggregated using
normalized reliability weights:

\begin{equation}
F_t^{\mathrm{fused}}
=
\sum_{i=1}^{N}\omega_i f_t^{i},
\qquad
\omega_i
=
\frac{R_t^{i}}
{\sum_{j=1}^{N}R_t^{j}},
\label{eq:reliability_fusion}
\end{equation}

followed by channel-attention refinement~\cite{wang2020ecanet,dey2025enhanced}:

\begin{equation}
F_t^{\mathrm{final}}
=
\operatorname{CA}
\left(
F_t^{\mathrm{fused}},f_t
\right).
\label{eq:channel_refinement}
\end{equation}

The resulting representation $F_t^{\mathrm{final}}$ is shared by the
downstream detection, segmentation, mapping, prediction, and planning
branches.

\subsection{Temporal Trajectory Planner}

The Temporal Trajectory Planner operates on the reliability-refined BEV
representation produced by RCA and performs temporal sequence modeling,
interaction reasoning, and multi-modal trajectory decoding. Rather than
relying only on the current-frame representation, the planner exploits
historical BEV information to preserve motion continuity and capture
longer-term dependencies among the ego vehicle, surrounding agents, and
map elements.

\begin{figure}
    \centering
    \includegraphics[width=.9\columnwidth, keepaspectratio]{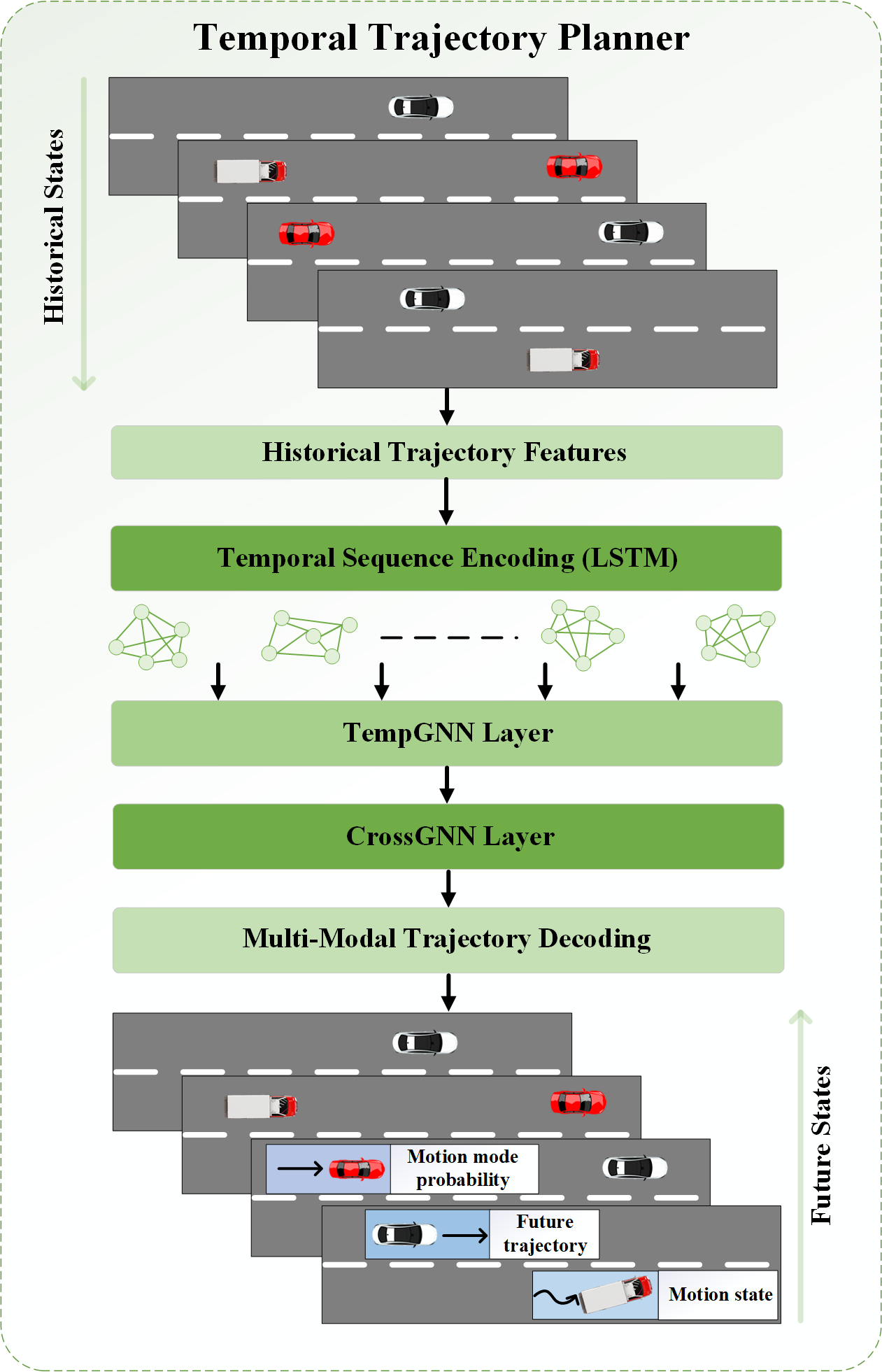}
    \caption{Overview of the proposed Temporal Trajectory Planner.
    Historical BEV-based motion representations are temporally encoded
    and subsequently refined through graph-based interaction reasoning.
    Agent-level temporal interactions are modeled by TempGNN, while
    CrossGNN incorporates interactions with map context. The resulting
    representation is decoded to predict future trajectories, motion
    states, and planning-related outputs.}
    \label{fig:temporal_planner}
\end{figure}

\subsubsection{Temporal Sequence Encoding}

For each motion mode $m$, a sequence of BEV-based motion embeddings
$\mathbf{Q}_{1:t}^{m}\in\mathbb{R}^{T\times C}$ is constructed from the
historical representations. A single-layer long short-term memory (LSTM)
network is employed to encode temporal dependencies:

\begin{equation}
\left(
\mathbf{H}_{t}^{m},
(\mathbf{h}_{t},\mathbf{c}_{t})
\right)
=
\operatorname{LSTM}
\left(
\mathbf{Q}_{t}^{m},
(\mathbf{h}_{t-1},\mathbf{c}_{t-1})
\right),
\label{eq:lstm}
\end{equation}

where $\mathbf{H}_{t}^{m}$ denotes the temporally aggregated motion
representation, while $\mathbf{h}_{t}$ and $\mathbf{c}_{t}$ are the
hidden and cell states, respectively. The hidden dimension is set to
$C=256$, consistent with the BEV embedding dimension. This recurrent
encoding preserves motion information across consecutive frames and
provides temporal context for subsequent interaction reasoning.

\subsubsection{Spatial Interaction Reasoning}

Temporal information alone is insufficient for trajectory prediction
because the future motion of an agent also depends on neighboring agents
and the surrounding road structure. We therefore refine the temporal
representation using graph-based interaction reasoning. Specifically,
agent-level interactions are first encoded by a temporal graph neural
network, followed by cross-interaction with map features:

\begin{equation}
\mathbf{Z}_{t}^{m}
=
\operatorname{CrossGNN}
\left(
\operatorname{TempGNN}
\left(
\mathbf{H}_{t}^{m},\mathbf{A}
\right),
\mathbf{M}_{\mathrm{map}}
\right),
\label{eq:interaction}
\end{equation}

where $\mathbf{A}$ denotes the agent-level contextual representation and
$\mathbf{M}_{\mathrm{map}}$ represents map features. The temporal graph
reasoning captures interactions among traffic participants, whereas the
cross-graph operation incorporates structural information from the road
environment. The resulting representation $\mathbf{Z}_{t}^{m}$ therefore
combines temporal motion context with agent-to-agent and agent-to-map
dependencies.

\subsubsection{Multi-Modal Trajectory Decoding}

The interaction-enhanced representation is decoded through lightweight
prediction heads to estimate the probability of each motion mode, future
waypoints, and motion state:

\begin{equation}
\hat{\mathbf{p}}^{m}
=
\operatorname{softmax}
\left(
\Psi_{\mathrm{cls}}
\left(
\mathbf{Z}_{t}^{m}
\right)
\right),
\label{eq:trajectory_cls}
\end{equation}

\begin{equation}
\hat{\mathbf{y}}^{m}
=
\Psi_{\mathrm{reg}}
\left(
\mathbf{Z}_{t}^{m}
\right),
\qquad
\hat{\mathbf{s}}^{m}
=
\Psi_{\mathrm{stat}}
\left(
\mathbf{Z}_{t}^{m}
\right),
\label{eq:trajectory_decode}
\end{equation}

where $\Psi_{\mathrm{cls}}$, $\Psi_{\mathrm{reg}}$, and
$\Psi_{\mathrm{stat}}$ denote the maneuver-classification, waypoint-
regression, and motion-state prediction heads, respectively. 

\subsubsection{Planning Objective}

The Temporal Trajectory Planner is optimized using a multi-task planning
objective that combines maneuver classification, trajectory regression,
and motion-state prediction:

\begin{equation}
\begin{split}
\mathcal{L}_{\mathrm{plan}}
={}&
\lambda_{\mathrm{cls}}
\mathcal{L}_{\mathrm{CE}}
\left(
\hat{\mathbf{p}}^{m},\mathbf{p}^{m}
\right)
\\
&+
\lambda_{\mathrm{reg}}
\left\|
\hat{\mathbf{y}}^{m}-\mathbf{y}^{m}
\right\|_{1}
+
\lambda_{\mathrm{stat}}
\mathcal{L}_{\mathrm{CE}}
\left(
\hat{\mathbf{s}}^{m},\mathbf{s}^{m}
\right),
\end{split}
\label{eq:planning_loss}
\end{equation}

where $\mathbf{p}^{m}$, $\mathbf{y}^{m}$, and $\mathbf{s}^{m}$ denote the
ground-truth maneuver label, future trajectory, and motion state,
respectively.

\subsection{Semantic and Motion-Guided BEV Refinement}

\begin{figure}
    \centering
    \includegraphics[width=.9\columnwidth, keepaspectratio]{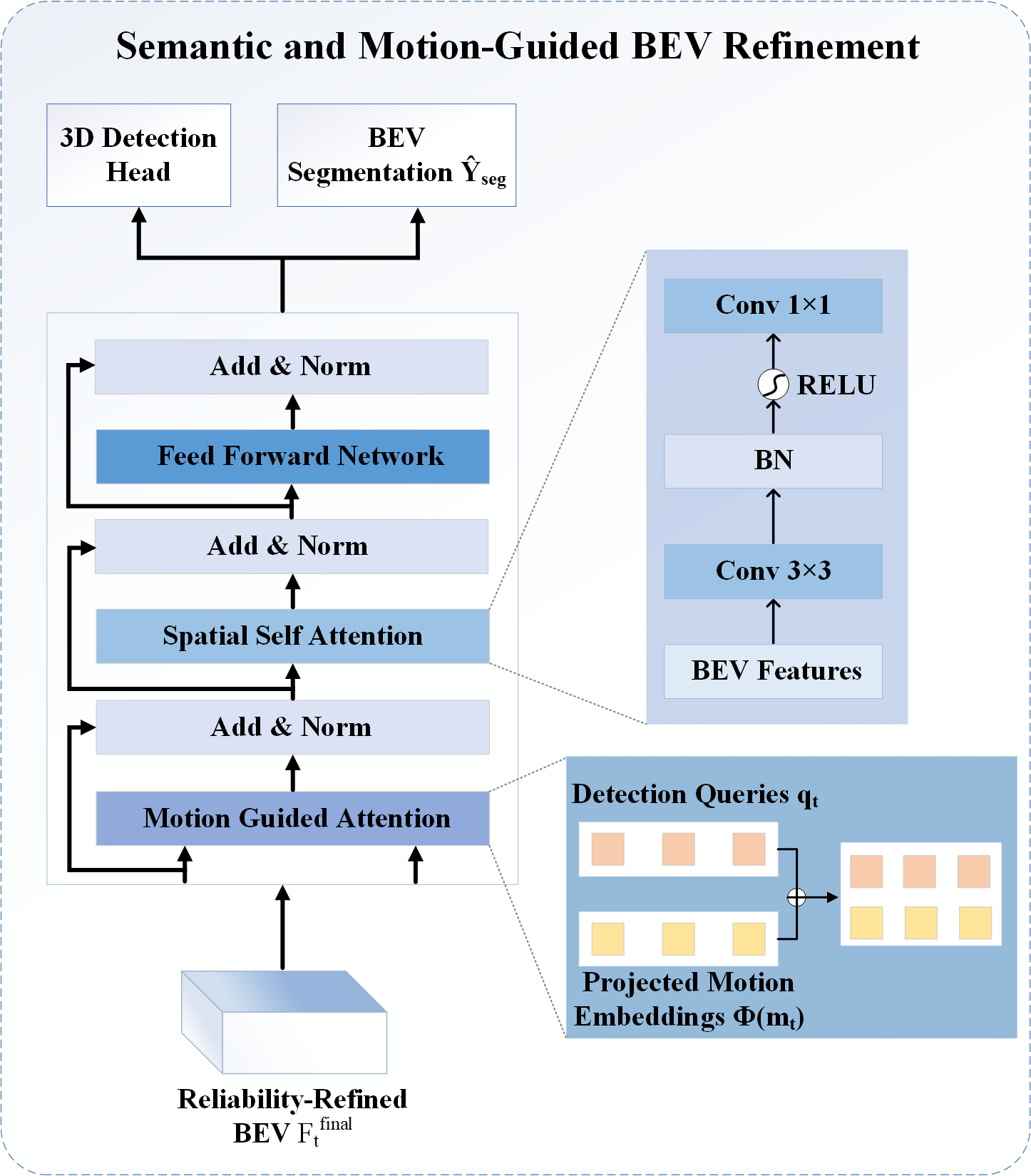}
    \caption{Overview of the proposed semantic and motion-guided BEV
    refinement module. The reliability-refined BEV representation
    $F_t^{\mathrm{final}}$ is processed using motion-guided attention,
    spatial self-attention, and feed-forward refinement. Detection queries
    are conditioned on projected motion embeddings, while spatially refined
    BEV features are further processed by a convolutional segmentation
    branch. The resulting representations are used for 3D object detection
    and BEV semantic segmentation.}
    \label{fig:bev_refinement}
\end{figure}

To further enrich the reliability-refined BEV representation with
object-level motion information and dense scene semantics, RCT-AD
introduces a semantic and motion-guided BEV refinement module, as
illustrated in Fig.~\ref{fig:bev_refinement}. The module takes the
reliability-refined representation $F_t^{\mathrm{final}}$ produced by
RCA as input and jointly supports 3D object detection and BEV semantic
segmentation.

The refinement process consists of motion-guided query enhancement,
spatial self-attention, and feed-forward feature transformation.
Detection queries are first conditioned on projected motion embeddings
through a motion-guided attention operation. The resulting
motion-enhanced representations are subsequently refined through spatial
self-attention to capture interactions among BEV elements. A feed-forward
network further transforms the refined features, with residual
connections and normalization applied after each attention or
transformation stage. The resulting representation provides a common
basis for the 3D detection head, while spatially refined BEV features
are also supplied to the semantic segmentation branch to preserve
scene-level structural information.

\subsubsection{Motion-Guided Detection}

To incorporate temporal motion information into object detection, the
motion embedding $\mathbf{m}_t$ is first projected into the
detection-query space through a learnable transformation $\phi(\cdot)$.
The projected motion representation is combined with the current
detection query $\mathbf{q}_t$ before attention-based refinement:

\begin{equation}
\mathbf{q}_{t}^{\prime}
=
\operatorname{Attn}
\left(
\mathbf{q}_{t}
+
\phi(\mathbf{m}_{t})
\right),
\label{eq:motion_detection}
\end{equation}

where $\phi(\cdot)$ aligns the motion embedding with the dimensionality
of the detection queries, and $\operatorname{Attn}(\cdot)$ denotes the
motion-guided attention operation.

As illustrated in Fig.~\ref{fig:bev_refinement}, the motion-enhanced
queries are followed by residual normalization and spatial self-attention,
allowing object queries to exchange spatial information within the BEV
representation. A feed-forward network then further transforms the
features, and an additional residual normalization stage produces the
final representation used by the 3D detection head. This hierarchical
refinement enables the detection queries to jointly encode temporal
motion cues and spatial interaction information.

\subsubsection{Semantic BEV Segmentation}

In parallel with object-level detection, spatially refined BEV features
are provided to a lightweight semantic segmentation branch to preserve
dense scene-level structure. Let $F_t^{\mathrm{spa}}$ denote the BEV
representation obtained after spatial self-attention in the refinement
module. The semantic prediction is formulated as

\begin{equation}
\hat{\mathbf{Y}}_{\mathrm{seg}}
=
\operatorname{Conv}_{1\times1}
\left(
\operatorname{ReLU}
\left(
\operatorname{BN}
\left(
\operatorname{Conv}_{3\times3}
\left(
F_t^{\mathrm{spa}}
\right)
\right)
\right)
\right),
\label{eq:semantic_segmentation}
\end{equation}

where $\hat{\mathbf{Y}}_{\mathrm{seg}}$ denotes the predicted BEV
semantic map. The $3\times3$ convolution extracts local spatial context,
followed by batch normalization and nonlinear activation, while the
$1\times1$ convolution projects the refined representation into the
semantic output space. This dense semantic supervision encourages the
shared BEV representation to preserve structural information about
drivable regions, lanes, vehicles, pedestrians, and other relevant
scene elements, thereby providing complementary context for perception,
motion prediction, and trajectory planning.

\subsection{Unified Multi-Task Joint Optimization}

RCT-AD is trained in an end-to-end multi-task manner so that the shared
BEV representation receives complementary supervision from perception,
mapping, temporal reasoning, and trajectory planning. The overall
objective is defined as

\begin{equation}
\begin{split}
\mathcal{L}_{\mathrm{total}}
={}&
\lambda_{\mathrm{det}}\mathcal{L}_{\mathrm{det}}
+
\lambda_{\mathrm{seg}}\mathcal{L}_{\mathrm{seg}}
+
\lambda_{\mathrm{plan}}\mathcal{L}_{\mathrm{plan}}
\\
&+
\lambda_{\mathrm{map}}\mathcal{L}_{\mathrm{map}}
+
\lambda_{\mathrm{mem}}\mathcal{L}_{\mathrm{mem}},
\end{split}
\label{eq:total_loss}
\end{equation}

where $\mathcal{L}_{\mathrm{det}}$ denotes the object-detection loss,
$\mathcal{L}_{\mathrm{seg}}$ is the BEV semantic-segmentation loss,
$\mathcal{L}_{\mathrm{plan}}$ is the trajectory-planning objective defined
in Eq.~\eqref{eq:planning_loss}, $\mathcal{L}_{\mathrm{map}}$ supervises
the vectorized map representation, and $\mathcal{L}_{\mathrm{mem}}$
encourages temporal consistency of the reliability-refined
representations.

The corresponding loss weights are set to
$\lambda_{\mathrm{det}}=1.0$,
$\lambda_{\mathrm{seg}}=1.0$,
$\lambda_{\mathrm{plan}}=2.0$,
$\lambda_{\mathrm{map}}=0.5$, and
$\lambda_{\mathrm{mem}}=0.2$.

\section{Experiments}

This section evaluates RCT-AD from four perspectives: overall performance
on perception, motion prediction, and trajectory planning; the contribution
of individual framework components; sensitivity to the reliability-aware
design; and computational efficiency. All quantitative experiments are
conducted on the nuScenes validation set following commonly adopted
open-loop evaluation protocols.

\subsection{Dataset and Evaluation Protocol}

We evaluate RCT-AD on the nuScenes benchmark~\cite{caesar2020nuscenes},
which contains 1,000 driving scenes collected under diverse urban
conditions. Each scene provides synchronized multi-view camera streams
together with 3D object annotations, semantic information, vehicle poses,
and map data. Following established end-to-end autonomous-driving
protocols~\cite{jiang2023vad,zhang2025bridging}, all quantitative results are reported on
the nuScenes validation set.

The evaluation covers perception, motion prediction, tracking, and
trajectory planning. For trajectory planning, we report the $L_2$
displacement error and collision rate at prediction horizons of 1, 2,
and 3~s. Motion prediction is evaluated using average displacement error
(ADE), final displacement error (FDE), end-point accuracy (EPA), and
miss rate (MR). Detection performance is measured using mean average
precision (mAP) and the nuScenes Detection Score (NDS), while BEV
semantic segmentation is evaluated using mean intersection-over-union
(mIoU). For multi-object tracking, we report average multi-object
tracking accuracy (AMOTA), average multi-object tracking precision
(AMOTP), recall, and identity switches (IDS).

Unless otherwise specified, the reported results follow the evaluation
settings adopted by the corresponding baseline methods. Comparisons are
grouped according to backbone capacity where appropriate to reduce the
influence of differences in feature-extraction complexity. 

\subsection{Implementation Details}

RCT-AD predicts a 3-s future trajectory for the ego vehicle and a
6-s trajectory for surrounding agents, corresponding to
$T_{\mathrm{plan}}=6$ and $T_{\mathrm{mot}}=12$ future steps,
respectively. For temporal reasoning, we use the previous
$T_{\mathrm{m2m}}=6$ frames for motion prediction and
$T_{\mathrm{p2p}}=3$ frames for trajectory planning. The
reliability-aware temporal memory maintains up to $K=6$ historical
BEV representations.

We evaluate two model configurations. \textbf{RCT-AD-T} employs a
ResNet-50 backbone with an input resolution of $256\times704$, whereas
\textbf{RCT-AD-L} uses ResNet-101 with an input resolution of
$512\times1408$. Both configurations share the same reliability-aware
temporal reasoning, semantic refinement, and trajectory-planning
architecture.

The models are optimized using AdamW~\cite{loshchilov2019decoupled} with an initial learning rate of
$1\times10^{-4}$, a weight decay of $1\times10^{-3}$, and a cosine
learning-rate schedule. Training is performed for 50 epochs with a
batch size of 8. We adopt a two-stage training strategy: the perception
components are first initialized through perception-oriented training,
after which the complete architecture is jointly optimized using the
multi-task objective in Eq.~\eqref{eq:total_loss}. Unless otherwise
stated, the reliability threshold is set to $\tau=0.85$, and the
reliability coefficients are
$(\alpha_1,\alpha_2,\alpha_3,\alpha_4,\alpha_5)
=(0.25,0.30,0.15,0.20,0.10)$.

\subsection{Comparison with State-of-the-Art Methods}
\subsubsection{Motion Planning}

\begin{table*}[!t]
\centering
\caption{Motion planning comparison on the nuScenes validation set.
$\dagger$ indicates evaluation using the official checkpoint.}
\label{tab:planning}
\small
\setlength{\tabcolsep}{5pt}
\begin{tabular}{lccccccccc}
\toprule
\multirow{2}{*}{Method} &
\multirow{2}{*}{Backbone} &
\multicolumn{4}{c}{$L_2$ (m) $\downarrow$} &
\multicolumn{4}{c}{Collision Rate (\%) $\downarrow$} \\
\cmidrule(lr){3-6}
\cmidrule(lr){7-10}
&
&
1s & 2s & 3s & Avg &
1s & 2s & 3s & Avg \\
\midrule

GenAD~\cite{zheng2024genad}
& R50
& 0.36 & 0.83 & 1.55 & 0.91
& 0.06 & 0.23 & 1.00 & 0.43 \\

VAD~\cite{jiang2023vad}$^{\dagger}$
& R50
& 0.41 & 0.70 & 1.05 & 0.72
& 0.03 & 0.19 & 0.43 & 0.21 \\

SparseDrive~\cite{sun2025sparsedrive}$^{\dagger}$
& R50
& 0.30 & 0.58 & 0.95 & 0.61
& 0.01 & 0.05 & 0.23 & 0.10 \\

\textbf{RCT-AD-T (Ours)}
& R50
& 0.29 & 0.57 & 0.93 & \textbf{0.59}
& 0.01 & 0.04 & 0.18 & \textbf{0.07} \\

\midrule

UniAD~\cite{hu2023planning}$^{\dagger}$
& R101
& 0.45 & 0.70 & 1.04 & 0.73
& 0.05 & 0.17 & 0.71 & 0.31 \\

BridgeAD~\cite{zhang2025bridging}
& R101
& 0.28 & 0.55 & 0.92 & 0.58
& 0.00 & 0.04 & 0.20 & 0.08 \\

\textbf{RCT-AD-L (Ours)}
& R101
& \textbf{0.23} & \textbf{0.52} & \textbf{0.89} & \textbf{0.54}
& 0.01 & \textbf{0.03} & \textbf{0.15} & \textbf{0.06} \\

\bottomrule
\end{tabular}

\vspace{1mm}
\begin{minipage}{0.98\textwidth}
\footnotesize
\textit{Note:} UniAD~\cite{hu2023planning} is measured on an NVIDIA Tesla A100 GPU.
All other methods are evaluated on an NVIDIA RTX A6000 GPU with
a batch size of 1.
\end{minipage}
\end{table*}

Table~\ref{tab:planning} compares RCT-AD with representative
end-to-end autonomous-driving methods using $L_2$ displacement error
and collision rate over 1--3~s planning horizons. Under the ResNet-50
setting, RCT-AD-T achieves an average $L_2$ error of 0.59~m and an
average collision rate of 0.07\%, compared with 0.61~m and 0.10\%
for SparseDrive~\cite{sun2025sparsedrive}. With the ResNet-101 backbone, RCT-AD-L further reduces
the average $L_2$ error to 0.54~m and the average collision rate to
0.06\%, compared with 0.58~m and 0.08\% for BridgeAD~\cite{zhang2025bridging}. The improvements
are also maintained at the 3-s horizon, where RCT-AD-L obtains an
$L_2$ error of 0.89~m and a collision rate of 0.15\%. These results
indicate that reliability-refined temporal representations provide
consistent benefits for trajectory planning across both backbone
configurations.

\subsubsection{Motion Prediction}

\begin{table*}[!t]
\centering
\caption{Motion prediction comparison on the nuScenes validation set.
$\dagger$ indicates evaluation using the official checkpoint.
Results are reported for the two principal categories, cars and pedestrians.}
\label{tab:motion_prediction}
\small
\setlength{\tabcolsep}{5pt}
\begin{tabular}{lcccccccccc}
\toprule
\multirow{2}{*}{Method} &
\multirow{2}{*}{Backbone} &
\multicolumn{2}{c}{ADE (m) $\downarrow$} &
\multicolumn{2}{c}{FDE (m) $\downarrow$} &
\multicolumn{2}{c}{EPA $\uparrow$} &
\multicolumn{2}{c}{MR (\%) $\downarrow$} \\
\cmidrule(lr){3-4}
\cmidrule(lr){5-6}
\cmidrule(lr){7-8}
\cmidrule(lr){9-10}
&
&
Car & Ped &
Car & Ped &
Car & Ped &
Car & Ped \\
\midrule

SparseDrive~\cite{sun2025sparsedrive}$^{\dagger}$
& R50
& 0.62 & 0.72
& 0.99 & 1.07
& 0.48 & 0.41
& 0.14 & 0.14 \\

BridgeAD-S~\cite{zhang2025bridging}
& R50
& 0.62 & 0.70
& 0.98 & 0.99
& 0.50 & 0.44
& 0.13 & 0.13 \\

\textbf{RCT-AD-T (Ours)}
& R50
& \textbf{0.61} & \textbf{0.67}
& \textbf{0.95} & \textbf{0.98}
& \textbf{0.51} & \textbf{0.46}
& \textbf{0.12} & \textbf{0.11} \\

\midrule

ViP3D~\cite{gu2023vip3d}
& R101
& 2.05 & --
& 2.84 & --
& 0.23 & --
& 0.25 & -- \\

UniAD~\cite{hu2023planning}$^{\dagger}$
& R101
& 0.71 & 0.78
& 1.02 & 1.05
& 0.46 & 0.35
& 0.15 & 0.12 \\

BridgeAD-B~\cite{zhang2025bridging}
& R101
& 0.60 & 0.70
& 0.96 & 0.98
& 0.52 & 0.45
& 0.13 & 0.12 \\

\textbf{RCT-AD-L (Ours)}
& R101
& \textbf{0.58} & \textbf{0.66}
& \textbf{0.94} & \textbf{0.98}
& \textbf{0.56} & \textbf{0.47}
& \textbf{0.12} & \textbf{0.12} \\

\bottomrule
\end{tabular}
\end{table*}

Table~\ref{tab:motion_prediction} compares motion-prediction performance
for cars and pedestrians using ADE, FDE, EPA, and MR. Under the
ResNet-50 setting, RCT-AD-T achieves car/pedestrian ADE values of
0.61/0.67~m and FDE values of 0.95/0.98~m, while obtaining EPA values
of 0.51/0.46 and MR values of 0.12/0.11. These results improve upon
SparseDrive~\cite{sun2025sparsedrive} and BridgeAD~\cite{zhang2025bridging}-S across most reported metrics.

With the ResNet-101 backbone, RCT-AD-L further improves the
car/pedestrian ADE to 0.58/0.66~m and achieves FDE values of
0.94/0.98~m. It also obtains EPA values of 0.56/0.47 while maintaining
MR values of 0.12/0.12. The results indicate that temporal interaction
reasoning over the reliability-refined BEV representation provides
consistent benefits for future-motion prediction across both backbone
configurations.

\subsubsection{3D Object Detection}

\begin{table*}[!t]
\centering
\caption{3D object detection comparison on the nuScenes validation set.
$\dagger$ indicates evaluation using the official checkpoint.}
\label{tab:detection}
\scriptsize
\setlength{\tabcolsep}{3.5pt}

\resizebox{\textwidth}{!}{%
\begin{tabular}{lccccccccccccc}
\toprule
Method &
Backbone &
mAP $\uparrow$ &
NDS $\uparrow$ &
Car $\uparrow$ &
Truck $\uparrow$ &
Bus $\uparrow$ &
Trailer $\uparrow$ &
\shortstack{C.\\vehicle} $\uparrow$ &
Ped $\uparrow$ &
\shortstack{Motor\\cycle} $\uparrow$ &
Bicycle $\uparrow$ &
Barrier $\uparrow$ &
\shortstack{T.\\Cone} $\uparrow$ \\
\midrule

VAD~\cite{jiang2023vad}
& R50
& 27.3 & 39.7
& 44.5 & 18.6 & 32.1 & -- & --
& 38.7 & 23.1 & 41.4 & 27.4 & 47.2 \\

SparseDrive~\cite{sun2025sparsedrive}$^{\dagger}$
& R50
& 41.8 & 52.5
& 59.8 & 44.2 & 49.1 & -- & --
& 52.7 & 51.4 & 52.3 & 47.4 & 61.1 \\

BridgeAD-S~\cite{zhang2025bridging}
& R50
& 42.3 & 53.4
& 64.3 & 44.8 & 49.2 & -- & --
& 53.3 & 51.7 & 50.6 & 46.7 & 62.4 \\

\textbf{RCT-AD-T (Ours)}
& R50
& \textbf{44.2} & \textbf{55.6}
& \textbf{65.7} & \textbf{49.1} & \textbf{49.7} & -- & --
& \textbf{58.1} & \textbf{52.6} & \textbf{53.2}
& \textbf{47.1} & \textbf{66.9} \\

\midrule

BEVFormer~\cite{li2022bevformer}
& R101
& 41.6 & 51.7
& 59.6 & 44.4 & 48.8 & -- & --
& 51.9 & 50.8 & 51.2 & 47.5 & 61.8 \\

UniAD~\cite{hu2023planning}
& R101
& 38.0 & 49.8
& 53.4 & 39.5 & 42.9 & -- & --
& 47.2 & 45.3 & 50.1 & 46.3 & 55.3 \\

BridgeAD-B~\cite{zhang2025bridging}
& R101
& 50.7 & 59.4
& 76.2 & 47.2 & 49.8 & -- & --
& 67.4 & 59.1 & 61.9 & 68.3 & 77.1 \\

\textbf{RCT-AD-L (Ours)}
& R101
& \textbf{52.9} & \textbf{61.5}
& \textbf{79.1} & \textbf{47.9} & \textbf{52.2} & -- & --
& \textbf{72.5} & \textbf{63.6} & \textbf{62.2}
& \textbf{69.7} & \textbf{81.8} \\

\bottomrule
\end{tabular}%
}
\end{table*}

Table~\ref{tab:detection} compares the 3D object detection performance
of RCT-AD with representative BEV-based and end-to-end autonomous-driving
methods. Under the ResNet-50 setting, RCT-AD-T achieves 44.2 mAP and
55.6 NDS, improving upon BridgeAD~\cite{zhang2025bridging}-S, which obtains 42.3 mAP and
53.4 NDS. RCT-AD-T also achieves the strongest performance among the
compared ResNet-50 methods across most reported object categories.

With the ResNet-101 backbone, RCT-AD-L achieves 52.9 mAP and 61.5 NDS,
compared with 50.7 mAP and 59.4 NDS for BridgeAD~\cite{zhang2025bridging}-B. Improvements are
also observed across several object categories, including cars,
pedestrians, motorcycles, bicycles, barriers, and traffic cones. These
results indicate that reliability-aware temporal modeling and
motion-guided refinement provide complementary benefits to the shared
BEV representation used for object perception.

\subsubsection{BEV Semantic Segmentation}

\begin{table*}[!t]
\centering
\caption{BEV semantic segmentation comparison on the nuScenes validation set.}
\label{tab:segmentation}
\small
\setlength{\tabcolsep}{5pt}

\begin{tabular}{lccc}
\toprule
Method & Backbone & Input Size & mIoU $\uparrow$ \\
\midrule

VPN~\cite{pan2020crossview}
& R50
& $256\times704$
& 43.8 \\

LSS~\cite{philion2020lift}
& R50
& $256\times704$
& 45.0 \\

\textbf{RCT-AD-T (Ours)}
& R50
& $256\times704$
& \textbf{49.8} \\

\midrule

BEVFormer~\cite{li2022bevformer}
& R101
& $900\times1600$
& 47.3 \\

Simple-BEV~\cite{harley2023simplebev}
& R101
& $448\times800$
& 47.4 \\

\textbf{RCT-AD-L (Ours)}
& R101
& $512\times1408$
& \textbf{52.3} \\

\bottomrule
\end{tabular}
\end{table*}

Table~\ref{tab:segmentation} compares the BEV semantic segmentation
performance of RCT-AD with representative camera-based BEV methods.
Under the ResNet-50 setting, RCT-AD-T achieves an mIoU of 49.8,
compared with 43.8 for VPN~\cite{pan2020crossview} and 45.0 for LSS~\cite{philion2020lift}. With the ResNet-101
backbone, RCT-AD-L further improves the mIoU to 52.3, exceeding the
47.3 and 47.4 obtained by BEVFormer~\cite{li2022bevformer} and Simple-BEV~\cite{harley2023simplebev}, respectively.
These results indicate that semantic supervision together with
reliability-aware temporal refinement improves the quality of the
shared BEV representation.

\subsubsection{Multi-Object Tracking}

\begin{table*}[!t]
\centering
\caption{Multi-object tracking performance comparison on the
nuScenes validation set.}
\label{tab:tracking}
\small
\setlength{\tabcolsep}{8pt}

\begin{tabular}{lccccc}
\toprule
Method &
Backbone &
AMOTA $\uparrow$ &
AMOTP $\downarrow$ &
Recall $\uparrow$ &
IDS $\downarrow$ \\
\midrule

ViP3D~\cite{gu2023vip3d}
& R50
& 0.217
& 1.625
& 0.363
& -- \\

SparseDrive~\cite{sun2025sparsedrive}
& R50
& 0.386
& 1.254
& 0.499
& 886 \\

BridgeAD-S~\cite{zhang2025bridging}
& R50
& 0.398
& 1.232
& --
& 639 \\

\textbf{RCT-AD-T (Ours)}
& R50
& \textbf{0.415}
& \textbf{1.188}
& \textbf{0.530}
& \textbf{578} \\

\midrule

UniAD~\cite{hu2023planning}
& R101
& 0.359
& 1.320
& 0.467
& 906 \\

BridgeAD-B~\cite{zhang2025bridging}
& R101
& 0.512
& 1.080
& --
& 544 \\

\textbf{RCT-AD-L (Ours)}
& R101
& \textbf{0.527}
& \textbf{1.004}
& \textbf{0.519}
& \textbf{503} \\

\bottomrule
\end{tabular}
\end{table*}

Table~\ref{tab:tracking} compares the multi-object tracking performance
of RCT-AD with representative end-to-end autonomous-driving methods.
Under the ResNet-50 setting, RCT-AD-T achieves an AMOTA of 0.415 and
an AMOTP of 1.188, together with a recall of 0.530 and 578 identity
switches. Compared with BridgeAD~\cite{zhang2025bridging}-S, RCT-AD-T improves AMOTA from
0.398 to 0.415 and reduces AMOTP from 1.232 to 1.188 and identity
switches from 639 to 578.

With the ResNet-101 backbone, RCT-AD-L achieves an AMOTA of 0.527
and an AMOTP of 1.004 while reducing the number of identity switches
to 503. Compared with BridgeAD~\cite{zhang2025bridging}-B, this corresponds to an improvement
from 0.512 to 0.527 in AMOTA, from 1.080 to 1.004 in AMOTP, and from
544 to 503 identity switches. These results provide additional evidence
that reliability-aware temporal reasoning improves the consistency of
object representations across consecutive observations.

\subsection{Ablation and Sensitivity Analysis}
\subsubsection{Component-Wise Ablation}

We first investigate the contribution of the three major components of
RCT-AD: Reliable Context Awareness (RCA), the Temporal Trajectory Planner
(TTP), and Semantic and Motion-Guided BEV Refinement (SMR).
Table~\ref{tab:component_ablation} reports the results obtained by
selectively enabling these components under the RCT-AD-L configuration.

Configurations 1--3 evaluate the contribution of the downstream temporal
reasoning and semantic-refinement modules while RCA remains enabled.
Configuration 4 removes RCA while retaining both TTP and SMR, thereby
providing a direct comparison with the complete model in Configuration 5.
This comparison isolates the contribution of reliability-aware temporal
feature selection and reconstruction from the remaining components.

\begin{table*}[t]
\centering
\caption{Component-wise ablation of RCT-AD on the nuScenes validation set
using the RCT-AD-L configuration.}
\label{tab:component_ablation}
\begin{tabular}{cccccccccc}
\toprule
ID & RCA & TTP & SMR &
mAP $\uparrow$ & NDS $\uparrow$ & mIoU $\uparrow$ &
ADE $\downarrow$ & FDE $\downarrow$ & EPA $\uparrow$ \\
\midrule
1 & \checkmark & $\times$ & $\times$
  & -- & -- & -- & 0.71 & 1.53 & 0.38 \\

2 & \checkmark & $\times$ & \checkmark
  & 48.4 & 53.0 & 49.7 & 0.71 & 1.53 & 0.38 \\

3 & \checkmark & \checkmark & $\times$
  & 49.6 & 59.7 & 51.1 & 0.67 & 1.04 & 0.46 \\

4 & $\times$ & \checkmark & \checkmark
  & 50.8 & 58.8 & 51.5 & 0.66 & 1.01 & 0.47 \\

5 & \checkmark & \checkmark & \checkmark
  & \textbf{52.9} & \textbf{61.5} & \textbf{52.3}
  & \textbf{0.62} & \textbf{0.96} & \textbf{0.51} \\
\bottomrule
\end{tabular}
\end{table*}

Comparing Configurations 4 and 5 shows the direct effect of RCA while
keeping the downstream architecture unchanged. Enabling RCA changes
mAP from 50.8 to 52.9, NDS from 58.8 to 61.5, and mIoU from 51.5 to
52.3, while also improving motion-prediction metrics. This comparison
indicates that explicitly evaluating and refining temporal observations
provides complementary benefits beyond the trajectory-reasoning and
semantic-refinement modules. Configurations 2 and 3 further show that
SMR primarily strengthens perception-oriented representations, whereas
TTP provides a larger contribution to temporal motion prediction.
The complete configuration achieves the strongest overall performance,
indicating that the three components are complementary.

\subsubsection{Analysis of Reliability-Aware Temporal Memory}

We further examine the interaction between reliability-aware feature
processing and temporal memory. As shown in
Table~\ref{tab:memory_ablation}, reliability-aware processing without
temporal memory achieves an NDS of 52.0, whereas temporal memory without
reliability-aware filtering obtains 48.1. Combining both mechanisms
increases NDS to 55.6. These results suggest that temporal memory alone
is insufficient when historical observations are accumulated without
explicit reliability assessment, while reliability-guided memory provides
a more effective mechanism for temporal information propagation. 

\begin{table*}[t]
\centering
\caption{Ablation of reliability-aware processing and temporal memory on
the nuScenes validation set. The experiment was conducted on RCT-AD-T.}
\label{tab:memory_ablation}
\begin{tabular}{ccc}
\toprule
Reliability-Aware Processing & Temporal Memory & NDS $\uparrow$ \\
\midrule
\checkmark & $\times$ & 52.0 \\
$\times$ & \checkmark & 48.1 \\
\checkmark & \checkmark & \textbf{55.6} \\
\bottomrule
\end{tabular}
\end{table*}

\subsubsection{Sensitivity to Reliability Threshold}

The reliability threshold $\tau$ controls the trade-off between retaining
current observations and invoking reliability-guided reconstruction.
Table~\ref{tab:threshold} evaluates three threshold values. Increasing
$\tau$ from 0.65 to 0.75 improves mAP from 39.2 to 42.9 and NDS from
48.6 to 51.1. The best performance is obtained at $\tau=0.85$, yielding
44.2 mAP and 55.6 NDS. A higher threshold applies stricter reliability
screening and therefore reduces the contribution of low-confidence
observations to temporal memory. Based on these results, we use
$\tau=0.85$ in all remaining experiments.

\begin{table*}[t]
\centering
\caption{Sensitivity to the reliability threshold $\tau$ on the
nuScenes validation set. The experiment was conducted on RCT-AD-T.}
\label{tab:threshold}
\begin{tabular}{ccc}
\toprule
$\tau$ & mAP $\uparrow$ & NDS $\uparrow$ \\
\midrule
0.65 & 39.2 & 48.6 \\
0.75 & 42.9 & 51.1 \\
0.85 & \textbf{44.2} & \textbf{55.6} \\
\bottomrule
\end{tabular}
\end{table*}

\subsubsection{Sensitivity to Reliability Coefficients}

We additionally investigate the sensitivity of the reliability score to
its two largest coefficients, $\alpha_1$ for spatial consistency and
$\alpha_2$ for detection confidence. Starting from the default setting
$(0.25,0.30,0.15,0.20,0.10)$, either $\alpha_1$ or $\alpha_2$ is
perturbed by $\pm0.10$. The remaining coefficients are proportionally
rescaled so that the simplex constraint
$\sum_{i=1}^{5}\alpha_i=1$ is preserved.

For a perturbed coefficient $\alpha_k'$, the remaining coefficients are
normalized according to

\begin{equation}
\alpha_j'
=
\alpha_j
\frac{1-\alpha_k'}{1-\alpha_k},
\qquad j\neq k .
\label{eq:coefficient_normalization}
\end{equation}

\begin{table*}[t]
\centering
\caption{Sensitivity analysis of the reliability coefficients on the
nuScenes validation set. All configurations satisfy
$\sum_{i=1}^{5}\alpha_i=1$. The experiment was conducted on RCT-AD-T.}
\label{tab:coefficient_sensitivity}
\begin{tabular}{lcccccc}
\toprule
Variant &
$\alpha_1$ (IoU) &
$\alpha_2$ (Conf.) &
$\alpha_3$ ($1-H$) &
$\alpha_4$ ($S$) &
$\alpha_5$ ($P$) &
NDS $\uparrow$ \\
\midrule
$-\alpha_1$ &
0.150 & 0.340 & 0.170 & 0.227 & 0.113 & 54.8 \\

$+\alpha_1$ &
0.350 & 0.260 & 0.130 & 0.173 & 0.087 & 54.9 \\

$-\alpha_2$ &
0.286 & 0.200 & 0.171 & 0.229 & 0.114 & 54.3 \\

$+\alpha_2$ &
0.214 & 0.400 & 0.129 & 0.171 & 0.086 & 55.0 \\

Default &
0.250 & 0.300 & 0.150 & 0.200 & 0.100 &
\textbf{55.6} \\
\bottomrule
\end{tabular}
\end{table*}

Table~\ref{tab:coefficient_sensitivity} shows that moderate perturbations
of the reliability coefficients result in relatively limited performance
variation, while the default configuration provides the strongest overall
NDS. This indicates that the reliability formulation is not dependent on
a single narrowly tuned coefficient setting. The comparison between
$\alpha_1$ and $\alpha_2$ perturbations further reveals the relative
influence of spatial consistency and detection confidence on the temporal
reliability estimate.

\subsection{Computational Efficiency}

In addition to predictive performance, we evaluate the computational
characteristics of RCT-AD in terms of GPU memory consumption, inference
latency, floating-point operations (GFLOPs), parameter count, and
throughput. Table~\ref{tab:efficiency} summarizes the comparison with
representative end-to-end autonomous-driving frameworks.

RCT-AD requires 99.56~M parameters and 379~GFLOPs, with an inference
latency of 138.6~ms and a throughput of 7.2~FPS under the reported
evaluation setting. Compared with SparseDrive~\cite{sun2025sparsedrive} and BridgeAD~\cite{zhang2025bridging}, RCT-AD
introduces a moderate increase in memory consumption while maintaining
competitive computational complexity and inference efficiency. In
particular, the proposed reliability-aware temporal reasoning does not
introduce an excessive computational burden relative to the performance
improvements observed across perception, prediction, and planning tasks.

It should be noted that the runtime reported for UniAD~\cite{hu2023planning} was obtained on
an NVIDIA A100 GPU, whereas SparseDrive~\cite{sun2025sparsedrive}, BridgeAD~\cite{zhang2025bridging}, and RCT-AD were
evaluated on an NVIDIA RTX A6000 with a batch size of 1. Therefore,
direct latency and throughput comparisons with UniAD~\cite{hu2023planning} should be
interpreted with caution.

\begin{table*}[t]
\centering
\caption{Computational efficiency comparison of representative
end-to-end autonomous-driving methods.}
\label{tab:efficiency}
\begin{tabular}{lccccc}
\toprule
Method &
Memory (GB) &
Inference Time (ms) &
GFLOPs &
Params (M) &
FPS \\
\midrule
UniAD~\cite{hu2023planning}        & 50.0 & 555.6 & 1709 & 125.0  & 1.8 \\
SparseDrive~\cite{sun2025sparsedrive}  & 15.2 & 163.9 & 192  & 85.9   & 6.1 \\
BridgeAD~\cite{zhang2025bridging}     & 17.6 & 157.2 & 485  & 129.4  & 6.4 \\
RCT-AD-L       & 19.5 & \textbf{138.6} & 379 & 99.56 & \textbf{7.2} \\
\bottomrule
\end{tabular}

\vspace{1mm}
\footnotesize
\textit{Note:} UniAD~\cite{hu2023planning} was evaluated on an NVIDIA A100 GPU. SparseDrive~\cite{sun2025sparsedrive},
BridgeAD~\cite{zhang2025bridging}, and RCT-AD were evaluated on an NVIDIA RTX A6000 with batch
size 1. Hardware-dependent runtime metrics should therefore be interpreted
accordingly.
\end{table*}

\subsection{Qualitative Analysis}

\begin{figure*}
    \centering
    \includegraphics[width=0.85\textwidth]{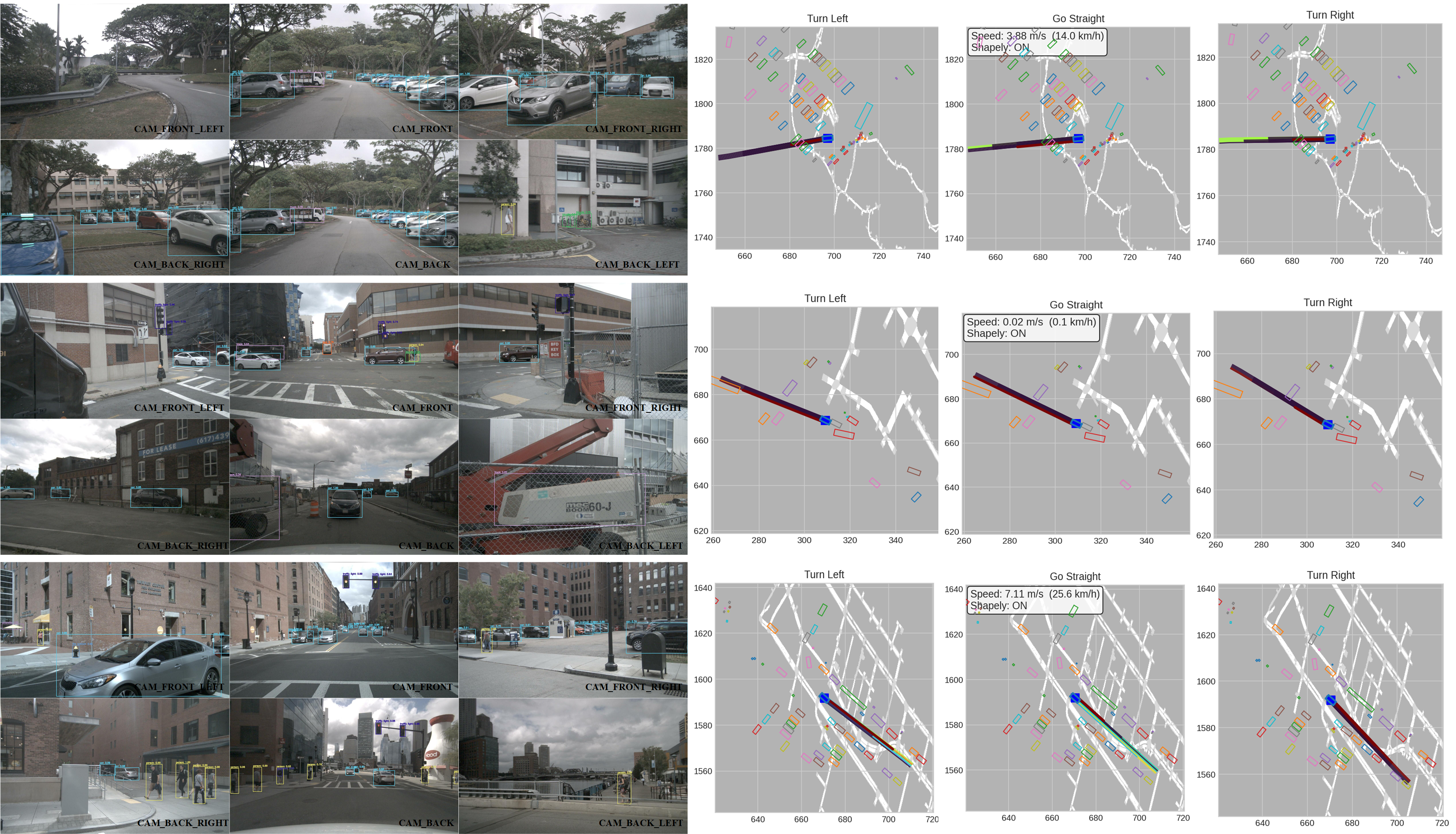}
    \caption{Qualitative results of RCT-AD on representative nuScenes scenes. Surround-view camera observations with detected traffic participants are shown on the left, while the corresponding BEV trajectory predictions for left-turn, straight, and right-turn maneuvers are shown on the right.}
    \label{fig:qualitative}
\end{figure*}

Figure~\ref{fig:qualitative} presents representative qualitative results
of RCT-AD on the nuScenes validation set. The visualization jointly
shows multi-view camera observations, object-level perception, BEV
semantic representations, and predicted trajectories, illustrating how
the different components interact within the unified framework.

The multi-view results show that object representations remain spatially
consistent across surrounding camera views and are subsequently projected
into the common BEV space. Through the RCA mechanism, temporal features
are selectively retained or reconstructed according to their estimated
reliability before being propagated to downstream tasks. The resulting
BEV representation provides a temporally refined basis for detection,
semantic understanding, and trajectory reasoning.

The BEV semantic outputs further illustrate the structural information
captured by the semantic-refinement branch, including drivable regions,
lane-related context, and surrounding traffic participants. Such
scene-level information complements object-centric representations and
provides additional contextual cues to the prediction and planning
modules.

The predicted trajectories illustrate the ability of the Temporal
Trajectory Planner to generate multiple motion hypotheses while
incorporating historical motion and interaction information. Across the
illustrated scenarios, the predicted trajectories remain consistent with
the surrounding scene geometry and observed traffic context. These
qualitative examples complement the quantitative results by illustrating
how reliability-aware temporal reasoning, semantic representation, and
trajectory prediction operate jointly within RCT-AD.

Overall, the quantitative and qualitative evaluations show that RCT-AD
provides consistent improvements across complementary autonomous-driving
tasks. The component ablations indicate that reliability-aware temporal
processing, trajectory reasoning, and semantic refinement contribute
complementary information, while the sensitivity analyses demonstrate
that the reliability mechanism remains effective across reasonable
threshold and coefficient settings. Together with the computational
analysis, these results support the use of reliability-aware temporal
BEV reasoning as a unified representation-learning strategy for
perception, prediction, and planning.

\section{Discussion and Future Work}

The experimental results indicate that explicitly modeling observation reliability can complement conventional temporal aggregation in BEV-based end-to-end autonomous driving. Rather than treating historical features as equally informative, RCT-AD evaluates their reliability before temporal propagation and uses high-reliability historical context to refine observations with lower estimated quality. The consistent improvements observed across detection, semantic segmentation, motion prediction, tracking, and trajectory planning suggest that reliability-aware temporal reasoning can improve multiple downstream tasks when they share a common BEV representation. In particular, the ablation results show that Reliable Context Awareness (RCA), temporal trajectory reasoning, and semantic and motion-guided refinement provide complementary benefits, indicating that the overall performance gain arises from their coordinated contribution to the shared representation rather than from a single task-specific component.

Several limitations remain. First, the current evaluation is conducted under the standard open-loop nuScenes protocol. Although the proposed reliability formulation incorporates spatial consistency, detection confidence, semantic uncertainty, temporal stability, and visual-quality cues, the experiments do not include a dedicated corruption benchmark in which the severity of adverse visual conditions is systematically controlled. Therefore, the reported results support the effectiveness of reliability-aware temporal representation learning under the evaluated scenarios, but they do not constitute a complete assessment of robustness to all forms of sensor degradation. Second, the reliability score is constructed using a predefined combination of complementary quality cues together with a fixed decision threshold. Although the sensitivity analyses examine the effects of the reliability threshold and coefficient configuration, a more adaptive estimator may generalize better across different scenes, sensors, and operating conditions. Third, the present study primarily focuses on camera-based BEV reasoning and open-loop evaluation, leaving interactive closed-loop behavior and broader multi-modal sensing settings for future investigation.

Future work will extend RCT-AD in several directions. One important direction is to evaluate the framework under explicitly controlled degradations, including blur, illumination changes, occlusion, image noise, and partial sensor failure, in order to characterize how reliability estimation changes as observation quality progressively deteriorates. Closed-loop evaluation in interactive environments, such as NeuroNCAP~\cite{ljungbergh2024neuroncap} and other reactive simulation platforms, will also be investigated to assess the effect of reliability-aware temporal reasoning when predicted actions influence subsequent observations. In addition, extending the framework to camera--LiDAR--radar fusion \cite{zhang2024transfusion, wu2023crossfuser} would enable reliability to be modeled not only across time but also across sensing modalities, allowing information from a reliable modality to compensate for degradation in another.

Another direction is to replace the manually specified reliability coefficients with a learnable or uncertainty-aware estimation mechanism. Probabilistic uncertainty modeling, confidence calibration, and dynamically conditioned reliability weighting could allow the contribution of individual reliability cues to adapt to the driving context. The temporal memory could also be extended with learnable retention and forgetting policies that jointly consider feature reliability, temporal age, scene dynamics, and task relevance. Finally, lightweight temporal reasoning, feature compression, and more efficient memory management will be explored to reduce computational overhead and improve scalability for longer temporal horizons. These extensions may contribute toward a more general framework for reliable spatiotemporal representation learning in autonomous systems.

\section{Conclusion}

This work presented RCT-AD, a reliability-aware temporal BEV reasoning framework for end-to-end autonomous driving. The framework addresses indiscriminate temporal feature propagation by explicitly estimating observation reliability before temporal information is incorporated into the shared BEV representation. Reliable Context Awareness combines complementary spatial, confidence, semantic-uncertainty, temporal-stability, and visual-quality cues to support reliability-guided feature selection, refinement, and memory fusion. On top of the resulting representation, the Temporal Trajectory Planner models long-term motion dependencies and agent interactions, while semantic and motion-guided refinement provides complementary supervision for perception and planning.

Experiments on the nuScenes validation set demonstrate consistent improvements across multiple tasks. RCT-AD-L achieves 52.9 mAP, 61.5 NDS, and 52.3 mIoU, together with an average planning displacement error of 0.54~m and an average collision rate of 0.06\%. The ablation and sensitivity analyses further show that reliability-aware temporal processing, trajectory reasoning, and semantic refinement provide complementary contributions to the overall framework. Overall, the results demonstrate that explicitly modeling observation reliability can improve the consistency of temporal BEV representations and support unified perception, prediction, and planning in end-to-end autonomous driving.

\section*{Acknowledgements}
The authors would like to thank the College of Computer Science,
Chongqing University, for providing the experimental facilities and
computational resources used in this study. This work was supported in
part by the National Natural Science Foundation of China under Grant
62262045, the Fundamental Research Funds for the Central Universities
under Grant 2023CDJYGRH-YB11, and the SUGON Industrial Control and
Security Center under Grant CUIT-SICSC-2025-03.

\section*{Competing interests}

The authors declare that they have no competing interests or financial conflicts to disclose.

\section*{Data availability}

The nuScenes dataset used in this study is publicly available from the
official nuScenes benchmark. No new dataset was created for this work. The source code will be made publicly available upon publication.

\section*{CRediT authorship contribution statement}

\noindent
\textbf{Argho Dey:}
Conceptualization, Methodology, Software, Validation,
Formal analysis, Investigation, Visualization,
Writing -- original draft, Project administration. \textbf{Yunfei Yin:}
Supervision, Writing -- review \& editing,
Funding acquisition, Project administration, Resources. \textbf{Swachha Ray:}
Data curation, Validation, Writing -- review \& editing. \textbf{Md Minhazul Islam:}
Software, Validation, Data curation. \textbf{Zheng Yuan:}
Formal analysis, Visualization, Writing -- review \& editing. \textbf{Sijing Xiong:}
Investigation, Data curation, Writing -- review \& editing. \textbf{Hongyu Liu:}
Software Implementation, Writing -- review \& editing. \textbf{Zhiqiu Huang:}
Formal analysis, Writing -- review \& editing.
\printcredits

\section*{Declaration of AI-assisted Manuscript Preparation}
AI-assisted writing tools supported manuscript organization, language revision and consistency checking. The authors are responsible for reviewing and verifying the text, references and scientific evidence and take full responsibility
for the article.

\bibliographystyle{cas-model2-names}
\bibliography{references}
\clearpage
\addtocounter{page}{-1}
\end{document}